\documentclass[journal]{IEEEtran}

\usepackage{times}
\usepackage{color}
\usepackage{verbatim}
\usepackage{amsmath, amssymb, amsthm}
\usepackage[bottom]{footmisc}
\usepackage{bbold}
\usepackage{graphicx}
\usepackage{algorithm, algorithmicx, algpseudocode}
\usepackage{verbatim}
\usepackage{cite}
\usepackage{url}
\usepackage{array}
\newcolumntype{P}[1]{>{\centering\arraybackslash}p{#1}}
\newcolumntype{M}[1]{>{\centering\arraybackslash}m{#1}}
\usepackage{multirow}
\newcommand{\noop}[1]{}

\newcommand{\RNum}[1]{\uppercase\expandafter{\romannumeral #1\relax}}

\newtheorem{proposition}{Proposition}

\newtheorem{remark}{Remark}

\begin{document}

\title{On Identification of Distribution Grids}


\author{\mbox{Omid~Ardakanian, 
        Vincent~W.S.~Wong,
        Roel~Dobbe,
        Steven~H.~Low,
        Alexandra~von~Meier,
        Claire~Tomlin,
        Ye~Yuan}
\thanks{Omid Ardakanian is with the Department of Computing Science, University of Alberta, Canada.
Vincent W.S. Wong is with the Department of Electrical and Computer Engineering, University of British Columbia, Canada.
Steven H. Low is with the Department of Computing and Mathematical Science, California Institute of Technology, USA.
Roel Dobbe, Alexandra von Meier, and Claire Tomlin are with the Department of Electrical Engineering and Computer Sciences, UC Berkeley, USA.
Ye Yuan is with the School of Automation, Huazhong University of Science and Technology, China.
For correspondence: \texttt{yye@hust.edu.cn}}}


\maketitle
\begin{abstract}
Large-scale integration of distributed energy resources into residential distribution feeders necessitates careful control of their operation through power flow analysis. While the knowledge of the distribution system model is crucial for this type of analysis, it is often unavailable or outdated. The recent introduction of synchrophasor technology in low-voltage distribution grids has created an unprecedented opportunity to learn this model from high-precision, time-synchronized measurements of voltage and current phasors at various locations. This paper focuses on joint estimation of model parameters (admittance values) and operational structure of a poly-phase distribution network from the available telemetry data via the lasso, a method for regression shrinkage and selection. We propose tractable convex programs capable of tackling the low rank structure of the distribution system and develop an online algorithm for early detection and localization of critical events that induce a change in the admittance matrix. The efficacy of these techniques is corroborated through power flow studies on four three-phase radial distribution systems serving real household demands.
\end{abstract}


\section{Introduction}

\IEEEPARstart{D}{istribution} grids are traditionally sized in a way 
that they will not be stressed even under severe loading conditions.
The emergence of demand side technologies and distributed energy resources (DER), 
such as solar panels, wind turbines, battery storage systems, and plug-in electric vehicles,
has led to an unprecedented amount of variability across multiple timescales
which cannot be simply managed using the traditional ``fit and forget'' philosophy.
This manifests the need for a novel operation paradigm which is centered around 
pervasive monitoring, real-time analytics and control at the distribution scale~\cite{Ardakanian16Book}.

In recent years, numerous smart grid technologies have been piloted 
to improve observability and controllability of distribution grids,
examples of which are distribution supervisory control and data acquisition (D-SCADA) system
and synchrophasor technology~\cite{naspi} -- 
inexpensive, high-fidelity micro-phasor measurement units ($\mu$PMUs) 
sampling voltage and current waveforms at high frequency~\cite{psl}.
The availability of telemetry data from multiple points across a distribution network 
makes valuable new applications possible,
such as event detection and classification, model validation, 
distributed generation (DG) characterization, state estimation,
equipment health monitoring, and phasor-based control~\cite{Ardakanian17a,Meier17}.
While the knowledge of the distribution network model is crucial 
for most of these applications,
such a model is often unavailable or outdated due to 
the continuous integration of distribution components and frequent reconfiguration of feeders.
To address this problem, this work focuses on 
joint estimation of distribution system model parameters 
and its topology from the available telemetry data.

The distribution system identification problem has received a lot of attention lately.
However, most research focuses on topology verification
which entails identifying the subset of distribution lines that are energized
using smart meter data or phasor measurements,
and little effort has been put into learning 
the impedance parameters of lines and transformers from the available data\footnote{Due to 
the imprecise and possibly correlated phasor measurements,
model parameter estimation will be nontrivial even if the true topology is known.}.
For example, the correlation between node voltage measurements is leveraged in~\cite{Bolognani13} 
to detect the grid topology via a sparse Markov random field.
A data-driven online algorithm is proposed in~\cite{Cavraro15} 
for detecting a switching event by comparing a trend vector built from $\mu$PMU data 
with a given library of signatures derived from the possible topology changes.
In~\cite{Sharon12}, the optimal placement of sensors 
in a distribution network is investigated 
in order to infer the status of switches 
from the measurements using the maximum likelihood method.
A mutual information-based algorithm is proposed in~\cite{Weng16} 
to identify the distribution topology
by building a graphical model that describes 
the probabilistic relationship among voltage measurements.
In~\cite{Deka16}, a graphical model learning algorithm is proposed
based on conditional independence tests for nodal voltage measurements.
Principal component analysis is employed in~\cite{Le14} 
to obtain a lower dimensional subspace of the available $\mu$PMU data
and project the original data onto this subspace by learning coefficients 
of the basis matrix using an adaptive training method.
An online event detection algorithm is then proposed 
to approximate phasor measurements using these coefficients,
issuing an alert whenever a significant approximation error is noticed.

The \emph{inverse power flow} (iPF) problem, originally defined in~\cite{Yuan16tr},
concerns recovering the admittance matrix of a power system 
from a sequence of voltage and current phasor measurements
corresponding to different steady states of the system.
In this paper, we study the iPF problem 
in the context of a poly-phase distribution system (loopy or radial)
where each node is equipped with a sensor (i.e., the full observability assumption).
Drawing on sparsity-based regularization techniques~\cite{Donoho03,Tropp05,venkat}, 
we present a tractable convex program to uniquely identify the admittance matrix 
of the distribution system when the identification problem is well-posed.
To tackle the low rank structure of a distribution network,
we develop a novel algorithm based on matrix decomposition 
which is capable of identifying a large submatrix of the admittance matrix.
Furthermore, we put forward an online algorithm 
for early detection and localization of critical events 
that change the admittance matrix.

This paper extends our prior work~\cite{Ardakanian17} in three main ways.
First, we propose the use of the adaptive lasso penalty  
to estimate large elements of the admittance matrix, attenuating the bias problem.
Second, we propose a regularization technique 
that leverages the approximate knowledge of the admittance matrix to enhance identification.
Third, we validate the efficacy of the proposed algorithms 
and study their sensitivity to the measurement error
through extensive simulations on IEEE 13, 34, 37, and 123-bus feeders 
serving real and synthetic loads.

The closest lines of work to ours are~\cite{Deka15,Yu17tr} 
which jointly address topology detection and model parameter estimation problems.
In~\cite{Deka15}, these problems are merely studied in a radial network setting
and the results are not extended to poly-phase and mesh systems.
In~\cite{Yu17tr}, noisy measurements of power injections and voltage phasors 
from $\mu$PMUs and smart meters are leveraged for the joint estimation of
line parameters and topology of a distribution system.
However, their approach cannot be used to detect faults or track topology changes.

\section{Problem Formulation}

This section describes the iPF problem in a poly-phase distribution system
and introduces a regularization technique for simultaneous estimation and variable selection
to efficiently solve this problem under certain assumptions.

\subsection{Preliminaries}
We denote the set of complex matrices
and the set of symmetric complex matrices by
$\mathbb{C}$ and $\mathbb{S}$, respectively,
the transpose of a matrix $A$ by $A^{\top}$,
its Hermitian (complex conjugate) transpose by $A^H$,
its pseudo-inverse by $A^\dagger$,
its Frobenius norm by $\|A\|_F$,
and the smallest number of linearly dependent columns of this matrix by $\textbf{Spark}\!\left(A\right)$.
All-zeros and all-ones vectors are denoted respectively by $\mathbf{0}$ and $\mathbf{1}$,
and 
the cardinality of a set $\mathcal{N}$ is denoted by $|\mathcal{N}|$.
Placing a caret over a letter indicates that it represents an estimated value.

A poly-phase power distribution system can be modeled 
by an undirected graph $\mathcal{G}=(\mathcal{N}, \mathcal{E})$ 
where $\mathcal{N}= \{1, 2, \ldots, N \}$ 
represents the set of nodes, 
and $\mathcal{E} \subseteq \mathcal{N} \times \mathcal{N}$ 
represents the set of energized lines, 
each connecting two distinct nodes.
We denote the phases of a node $n \in \mathcal{N}$ 
by $\mathcal{P}_n \subseteq \{a_n, b_n, c_n\}$
and the phases of a line $(m, n) \in \mathcal{E}$ 
connecting node $m$ to node $n$ by $\mathcal{P}_{mn} \subseteq \{a_{mn}, b_{mn}, c_{mn}\}$.
For node $n \in \mathcal{N}$ and phase $\phi \in \mathcal{P}_n$,
let $V^\phi_n \in \mathbb C$ be its line-to-ground voltage 
and $I^\phi_n \in \mathbb C$ be the injected current.
We represent the voltages and injected currents 
of different phases of node $n \in \mathcal{N}$
by vectors $V_n = \{V^\phi_n\}_{\phi \in \mathcal{P}_n}$ 
and $I_n = \{I^\phi_n\}_{\phi \in \mathcal{P}_n}$, respectively, 
and use the per-unit system to express the quantities.
We treat the voltage at the distribution substation as reference for phasor representation.
 
We model lines as $\pi$-equivalent components
and denote the phase impedance and shunt admittance matrices of line $(m,n)$
by $Z_{mn} \in \mathbb{C}^{|\mathcal{P}_{mn}|\times |\mathcal{P}_{mn}|}$
and $Y^s_{mn} \in \mathbb{C}^{|\mathcal{P}_{mn}|\times |\mathcal{P}_{mn}|}$,
respectively. 
Similarly, transformers are modeled as series components 
with an admittance matrix which depends on the type of connection.
Assembling the admittance matrices of all components, 
the admittance matrix can be constructed for the distribution system, 
denoted by $Y_{\text{bus}} \in \mathbb{S}^{\sum_{n \in \mathcal N} | \mathcal{P}_n | \times \sum_{n \in \mathcal N} | \mathcal{P}_n |}$,
which is a symmetric matrix that satisfies
$Y_{\text{bus}}\mathbf{1}=\mathbf{0}$ if shunt elements are neglected.
The bus admittance matrix relates the node voltages and injected currents 
according to Ohm's law:
\begin{equation}\label{eq:kl2}
\underbrace{\begin{bmatrix}
I_1 (k)\\
I_2 (k)\\
\vdots\\
I_N (k)
\end{bmatrix}}_{I_{\text{bus}}(k)}=
\underbrace{\begin{bmatrix} 
Y_{11} & Y_{12} & \ldots & Y_{1N} \\
Y_{12}^\top & Y_{22} & \ldots & Y_{2N} \\
\vdots & \vdots & \ddots & \vdots\\
Y_{1N}^\top & Y_{2N}^\top & \ldots & Y_{NN}
\end{bmatrix}}_{Y_{\text{bus}}}
\underbrace{\begin{bmatrix}
V_1 (k)\\
V_2 (k)\\
\vdots\\
V_N (k)
\end{bmatrix}}_{V_{\text{bus}}(k)},
\end{equation}
where $k=1,\ldots, K$ is the time index,
$V_{\text{bus}}(k), I_{\text{bus}}(k)\in\mathbb C^{\sum_{n \in \mathcal N} | \mathcal{P}_n |\times 1}$
are steady-state complex nodal voltages and injected currents at time $k$, 
each off-diagonal block of $Y_{\text{bus}}$ is a submatrix $Y_{mn} = -Z^{-1}_{mn}$ 
corresponding to the admittance of line $(m,n)$,
and each diagonal block is a submatrix
\[Y_{nn} = \sum_{m \in \{o | (o,n) \in \mathcal{E} \}} \left(\frac{1}{2}Y^s_{mn} + Z^{-1}_{mn}\right). \]

Rewriting (\ref{eq:kl2}) in vector form for $K$ time slots yields:
\begin{equation}\label{eq:kl}
\underbrace{\begin{bmatrix}
I_1 (1) & \ldots & I_1 (K)\\
I_2 (1) & \ldots & I_2 (K)\\
\vdots & \ddots & \vdots\\
I_N (1) & \ldots & I_N (K)
\end{bmatrix}}_{I_{\text{bus}}^K}\!\!=
\!Y_{\text{bus}}\!
\underbrace{\begin{bmatrix}
V_1 (1) & \ldots & V_1 (K)\\
V_2 (1) & \ldots & V_2 (K)\\
\vdots & \ddots & \vdots\\
V_N (1) & \ldots & V_N (K)\\
\end{bmatrix}}_{V_{\text{bus}}^K},
\end{equation}
where $V_{\text{bus}}^K$ and $I_{\text{bus}}^K$ 
collect nodal voltages and injected currents sampled at $K$ 
successive time slots, respectively.

The iPF problem that we study in this paper
concerns recovering the admittance matrix of a poly-phase distribution system, $Y_{\text{bus}}$,
from voltage and current phasor measurements of all nodes, $V_{\text{bus}}^K$ and $I_{\text{bus}}^K$.
In general, $V_{\text{bus}}^K$ is low rank in a power system\footnote{This is 
reported for transmission PMU data in~\cite{Le14} 
and is further supported by the experiments on real $\mu$PMU data obtained from~\cite{Meier14}.},
making the identification problem ill-posed.
In the following, we first study how the admittance matrix 
can be identified when the identification problem is well-posed. 
We discuss in Section~\ref{sec:lowrank} 
how a large part of the admittance matrix can be identified despite 
the low rank structure of $V_{\text{bus}}^K$.

\subsection{Sparsity-based Regularization}\label{sec:idgen}

This section presents a robust algorithm for recovering the admittance matrix 
of a distribution system from noisy sensor data\footnote{
Existing synchrophasor technology is capable of 
sampling voltage and current waveforms at 120~Hz or higher~\cite{psl}.
But even at a lower temporal resolution (e.g., one-minute),
enough data can be collected to identify the model 
of a distribution network comprised of several hundred nodes, before it changes.
Nevertheless, for convenience, we henceforth refer 
to the available sensor data as $\mu$PMU data.}
under two assumptions: 
(a) the identification problem is well-posed, and (b) all nodes can be monitored.

The admittance matrix can be identified 
by solving the following regression problem:
\begin{align}\label{eq:Yl2}
\widehat{Y}_{\text{bus}}=&\arg\!\min_{Y}
\left\| Y V_{\text{bus}}^K - I_{\text{bus}}^K \right\|_F \\
\nonumber &\text{subject to}\qquad Y \in \mathbb{S}^{N\times N}.
\end{align}
In practice, the sample size $K$ can be smaller than the number 
of unknown variables in the admittance matrix
and we seek for a sparse solution because $Y_\text{bus}$ 
encodes the topology of a distribution network\footnote{Distribution systems typically
have a radial operational structure since they are operated 
in such a way that power flows on a radial sub-graph at any particular time. 
Hence, most elements of the admittance matrix are zero.}.
Thus, we adopt sparsity-based regularization techniques to identify the admittance matrix.
We specifically enforce sparsity of $Y_\text{bus}$ 
by applying the $\textit{vec}$ operator, which converts a matrix into a column vector, 
to the objective function and constraining the $\ell_0$-norm of $\text{vec}(Y_\text{bus})$:

\begin{align}
\hat{Y}_{\text{bus}}=&\arg\!\min_{Y}
\left\| ((V_{\text{bus}}^K)^\top \otimes \mathbb{1}_N) \text{vec}(Y)\!-\! \text{vec}(I_{\text{bus}}^K) \right\|_2 \\
\nonumber &\text{subject to}\qquad Y \in \mathbb{S}^{N\times N},~~~\left\| \text{vec}(Y) \right\|_0 \leq \delta,
\end{align}
where 
$\otimes$ is the Kronecker product
and $\delta$ determines the degree of sparsity of $Y_{\text{bus}}$.
The cardinality constraint makes this problem NP-hard~\cite{Natarajan95}.

Exploiting the symmetric structure of $Y_{\text{bus}}$, 
we reduce the number of parameters that need to be estimated.
Consider a mapping
$f:\mathbb{C}^{N\times N} \rightarrow\mathbb{C}^{(N^2+N)/2\times 1}$
which collects the lower triangular elements of a complex matrix as illustrated below:
\begin{equation*}
f(A)=[a_{11}, a_{21}, a_{31},\ldots a_{N1}, a_{22}, a_{32},\ldots a_{N2}, \ldots a_{NN}]^\top,
\end{equation*}
where $a_{ij}$ is the element in the $i$th row and $j$th column of matrix $A$.
Observe that $f$ is a bijection for any $Y\in\mathbb{S}^{N\times N}$
and we have $\text{vec}(Y)= Q_Yf(Y)$, where $Q_Y\in\mathbb{R}^{N^2\times(N^2+N)/2}$
is a unique binary matrix that converts $f(Y)$ to $\text{vec}(Y)$.
Hence, the iPF problem can be reformulated as:
\begin{equation*}
f(\widehat{Y})\!\!=\!\!\arg\!\!\!\!\!\!\!\!\!\!\!\!\!\!\min_{x\in \mathbb{C}^{(N^2+N)/2 \times 1}} \!\left\|\!\left({V_{\text{bus}}^K}^\top\!\!\otimes \mathbb{1}_N\right)\!Q_Y x \!-\!\text{vec}(I_{\text{bus}}^K) \right\|_2^2\!+\! \lambda \left\| x \right\|_0\!,
\end{equation*}
where $\lambda$ is a suitable positive regularization parameter.
The above problem is non-convex and cannot be solved efficiently;
thus, we solve a convex relaxation of this problem 
known as the lasso~\cite{Tibshirani96}, 
hoping that the solutions coincide\footnote{In~\cite{Tropp05}, 
conditions are established for the solution of $\ell_1$ optimization to
coincide with the solution of $\ell_0$ optimization.}.
The penalized form of lasso can be written as:
\begin{equation}\label{eq:prob-lasso}
\!\!\min_{x\in \mathbb{C}^{(N^2+N)/2\times 1}} \!\left\| \underbrace{\left({V_{\text{bus}}^K}^\top\!\!\otimes \mathbb{1}_N\right)\!Q_Y}_{A}\!x \!-\!\underbrace{\text{vec}(I_{\text{bus}}^K)}_{b} \right\|_2^2 \!+\! \lambda \left\| x \right\|_1\!.
\end{equation}

The lasso continuously shrinks 
the elements of $Y_\text{bus}$ toward 0 as $\lambda$ increases,  
and some coefficients are shrunk to exact 0 if $\lambda$ is sufficiently large. 
Hence, selecting $\lambda$ is critical to the performance of the lasso
and cross-validation can be used for this purpose.
We note that \eqref{eq:prob-lasso} can be solved using a standard convex optimizer
as well as iterative algorithms~\cite{Daubechies04,Efron04,Friedman07},
which are more compelling in large distribution networks.


Once $f(\widehat{Y})$ is recovered, 
$\text{vec}(\widehat{Y})$ can be easily constructed:
\begin{equation}
\text{vec}(\widehat{Y}) = Q_Y f(\widehat{Y}).
\end{equation}
It is shown in~\cite{Yuan16tr} that the proposed technique 
can solve the identification problem when it is well-posed.

\subsection{Avoiding Unnecessary Bias}\label{sec:fullrank}

Despite significant statistical and computational advantages 
of the lasso for solving the iPF problem, 
it is not an `oracle procedure' and 
could result in suboptimal estimation in certain cases~\cite{Zou06}.
The lasso equally penalizes all elements of $Y_\text{bus}$, 
thereby producing biased estimates for the large elements.
Hence, it may fail to identify the true admittance matrix 
when the distribution system contains several switches and voltage regulators
which have much larger admittance than the distribution lines.


To avoid the unnecessary bias, we can assign data-dependent weights 
to different elements of $Y_\text{bus}$ in the $\ell_1$ penalty.
The two-stage algorithm, known as the adaptive lasso~\cite{Zou06},
applies less shrinkage whenever the true unknown variable is large.
Specifically, $f(Y)$ can be recovered from:
\begin{equation}\label{eq:adaptive}
\!\!\min_{x\in \mathbb{C}^{(N^2+N)/2 \times 1}} \!\left\| A x \!-\! b \right\|_2^2 \!\!+\! \lambda\!\sum_{i}\frac{|x_i|}{{|\hat{x}_i|}^\gamma},
\end{equation}
where $\gamma$ is a positive parameter 
and $\hat{x}_i$ is an initial estimator for $x_i$, 
e.g., the ordinary least squares (OLS) estimator defined:
\begin{align*}
\hat{x}=(A^\top A)^{-1}A^\top b.
\end{align*}
Rescaling the columns of 
$A=\left({V_{\text{bus}}^K}^\top\!\!\otimes \mathbb{1}_N\right)\!Q_Y$
with the corresponding weights, i.e., $|\hat{x}_i|^\gamma$,
reduces \eqref{eq:adaptive} to the standard lasso problem, 
and therefore, it can be solved using the same algorithms developed for the lasso.
Note that two-dimensional cross-validation 
is typically used to tune $(\lambda,\gamma)$.

In Section~\ref{sec:results}, 
we compare the lasso and the adaptive lasso penalties 
and show that the adaptive lasso outperforms the lasso 
in terms of identification accuracy in the test feeders.

\subsection{Exploiting Additional Structure}

Distribution circuits are upgraded and reconfigured
to meet the growing demand of a neighbourhood,
accommodate new technologies installed at customers' premises, and minimize losses.
These changes are seldom incorporated into the distribution system model;
thus, the available model is usually obsolete 
and cannot be relied on for diagnostics and control applications.
We now discuss how such an approximate model
can be leveraged to improve the identification accuracy.

We represent the available (and presumably inaccurate)
admittance matrix of a distribution system by 
$\widetilde{Y}\triangleq Y_\text{bus}+\Psi$ 
where $Y_\text{bus}$ is the true admittance matrix of the network, 
which we ultimately intend to find,
and $\Psi$ is an arbitrary error matrix that must be identified. 
We note that $\Psi$ is symmetric since both $\widetilde{Y}$ and $Y_\text{bus}$ are symmetric.
If all elements of $\Psi$ are small\footnote{Alternately, $\Psi$ might have a small number 
of nonzero elements that are not necessarily small, 
e.g., when the only unknown information is the status of switches which typically have large admittance values.
In such cases, the Frobenius norm of $\Psi$ in~\eqref{eq:partialinf} must be replaced with the $\ell_0$-norm of $\Psi$.}, 
the identification problem reduces to solving the following regularized least squares problem:
\begin{align}\label{eq:partialinf}
\widehat{Y}_{\text{bus}}=\widetilde{Y} - \arg\!\!\!\!\!\!\min_{\Psi\in \mathbb{S}^{N\times N}}&
\left\| (\widetilde{Y} - \Psi) V_{\text{bus}}^K - I_{\text{bus}}^K \right\|_F^2\!\!+ \!\lambda \left\| \Psi \right\|_F^2,
\end{align}
where $\lambda$ is a tuning parameter.

Exploiting the symmetric structure of $\Psi$ 
and adopting the technique outlined in Section~\ref{sec:idgen},
we can solve the following \emph{ridge regression} problem to identify $\Psi$:
\begin{equation}
f(\widehat{\Psi}) = \arg\!\!\!\!\!\!\!\!\!\!\!\min_{x\in \mathbb{C}^{(N^2+N)\times 1}} \left\|Ax - b\right\|^2_2 + \lambda \left\| x \right\|^2_2
\end{equation}
where $A = -\left({V_{\text{bus}}^K}^\top \otimes \mathbb{1}_N\right)Q_\Psi$ and $b = \text{vec}(I_{\text{bus}}^K - \widetilde{Y} V_{\text{bus}}^K)$.
This approach can be used to periodically update the distribution system model.

\section{Low Rank Structure of Distribution Systems}\label{sec:lowrank}

The voltage measurement matrix, $V_{\text{bus}}^K$, 
is low rank in most power distribution systems 
due to the interdependencies between nodal voltages.
This results in an ill-posed problem that cannot be solved 
to identify the admittance matrix in its entirety 
even if $K\gg N$.
To tackle this problem, 
we propose a novel identification algorithm based on
a particular partitioning of $V_{\text{bus}}^K$ into two matrices, 
one of which has full row rank;
this permits us to recover at least some part of the admittance matrix 
while the rest of it cannot be recovered.
The steps of this algorithm are described below.

\subsection{Similarity Transformation}

Let $R$ be the row rank of $V_{\text{bus}}^K$.
We partition $V_{\text{bus}}^K$ into two matrices via a similarity transformation of the matrix $Y_{\text{bus}}$:
\begin{equation}\label{eq:simtrans}
\underbrace{\mathcal T I_{\text{bus}}^K}_{\mathbb I} = \underbrace{(\mathcal T Y_{\text{bus}} \mathcal T^{-1})}_{\mathbb Y} (\underbrace{\mathcal T V_{\text{bus}}^K}_{\mathbb V}),
\end{equation}
where $\mathcal T$ is a $\sum_{n \in \mathcal N} | \mathcal{P}_n | \times \sum_{n \in \mathcal N} | \mathcal{P}_n |$ matrix 
that splits $V_{\text{bus}}^K$ into
an $R \times K$ matrix, denoted by $\mathbb V_2$, 
containing $R$ linearly independent rows of $V_{\text{bus}}^K$ and 
an $(\sum_{n \in \mathcal N} | \mathcal{P}_n | - R) \times K$ matrix, denoted by $\mathbb V_1$, 
containing other rows of $V_{\text{bus}}^K$ that are all in the row space of $\mathbb V_2$. 
Algorithm~\ref{alg:basis} describes the steps for
building these two submatrices from the available synchrophasor data.
Shuffling rows of $V_{\text{bus}}^K$ and $I_{\text{bus}}^K$ 
according to this transformation yields:
\[ \mathcal T V_{\text{bus}}^K = 
\begin{bmatrix} \mathbb V_1 \\ \mathbb V_2 \end{bmatrix}, ~\qquad
\mathcal T I_{\text{bus}}^K = 
\begin{bmatrix} \mathbb I_1 \\ \mathbb I_2 \end{bmatrix}.\]

Since $\mathbb V_1$ is in the row space of $\mathbb V_2$, 
we can estimate the basis $X$ such that $\mathbb V_1 = X \mathbb V_2$
from $\mu$PMU data by computing the pseudo-inverse of $\mathbb V_2$: $X = \mathbb V_1\mathbb V_2^{\dagger}$.
Note that the pseudo-inverse is well-defined here since $\mathbb V_2$ is full row rank.

\subsection{Recovering Parts of $Y_{\text{bus}}$}

We write~\eqref{eq:simtrans} as
\begin{equation}
\begin{bmatrix}
\mathbb I_1 \\ \mathbb I_2
\end{bmatrix}
= \underbrace{\begin{bmatrix} 
\mathbb Y_{1,1} & \mathbb Y_{1,2} \\
\mathbb Y_{1,2}^{\top} & \mathbb Y_{2,2}
\end{bmatrix}}_{\mathbb Y}
\begin{bmatrix}
X \mathbb V_2 \\ \mathbb V_2
\end{bmatrix} 
= \begin{bmatrix} 
\mathbb Y_{1,1} X + \mathbb Y_{1,2} \\
\mathbb Y_{1,2}^{\top} X + \mathbb Y_{2,2}
\end{bmatrix}
 \mathbb V_2,
\end{equation}
where $\{\mathbb{Y}_{i,j}\}_{i,j\in \{1,2\}}$ are four submatrices of $\mathbb{Y}$ 
obtained according to the decomposition of $\mathbb{V}$.
Note that $\mathbb V_2$ has full row rank. We have
\begin{align}
\label{eq:YX1} \mathbb I_1 &= \left(\mathbb Y_{1,1}X + \mathbb Y_{1,2}\right) \mathbb V_2,\\
\label{eq:YX2} \mathbb I_2 &= \left(\mathbb Y_{1,2}^{\top} X + \mathbb Y_{2,2}\right) \mathbb V_2.
\end{align}
Solving~\eqref{eq:YX1} for $\mathbb Y_{1,2}$ 
and substituting it into~\eqref{eq:YX2} yields:
\begin{equation}
-X^{\top} \mathbb Y_{1,1} X + \mathbb Y_{2,2} = C,
\end{equation}
in which $C=\mathbb I_2 \mathbb V_2^\dagger - (\mathbb V_2^\dagger)^{\top} \mathbb I_1^{\top} X$ 
can be computed from the $\mu$PMU data. 
Vectorizing both sides of the equation yields:
$$-\left(X^{\top} \otimes X^{\top} \right)\text{vec}(\mathbb Y_{1,1}) + \text{vec} (\mathbb Y_{2,2}) = \text{vec}(C).$$
This problem can be written in the following form to reduce the number of parameters 
that need to be estimated 
using bijection $f$ and matrix $Q$:
$$-\left(X^{\top} \otimes X^{\top} \right)Q_{Y_{11}}f(\mathbb Y_{1,1}) + Q_{Y_{22}}f(\mathbb Y_{2,2}) = \text{vec}(C).$$
Enforcing sparsity of the components of $Y_{\text{bus}}$,
it is possible to identify $\mathbb Y_{1,1}$ and $\mathbb Y_{2,2}$ from this optimization problem:
\begin{align}
\begin{bmatrix} f(\widehat{\mathbb{Y}}_{1,1}) \\ f(\widehat{\mathbb{Y}}_{2,2}) \end{bmatrix}=
&\arg\min_x \lambda\sum_{i} w_i|x_i| + \\
\nonumber &\left\| \begin{bmatrix} -(X^{\top}\!\!\otimes\!\!X^{\top})Q_{Y_{11}}, & \!\!\!\!Q_{Y_{22}} \end{bmatrix} x - \text{vec}(C) \right\|_2^2.
\end{align}
Depending on whether we use the lasso or the adaptive lasso penalty, 
$w_i$ is set to 1 or $1/|\hat{x}_i|^\gamma$ for some $\gamma>0$.


Once this problem is solved, 
$\mathbb Y_{1,2}$ can be identified from~\eqref{eq:YX2} using the method of least squares.
However, there is no guarantee that $\mathbb Y_{1,2}$ is estimated with sufficient accuracy
as the error introduced in the process of estimating $\mathbb{Y}_{2,2}$ propagates.
We show in Section~\ref{sec:results} that $\mathbb{Y}_{2,2}$
can be accurately estimated in all cases despite the low rank structure,
while $\mathbb Y_{1,1}$ and $\mathbb Y_{1,2}$ cannot be recovered with sufficient accuracy.

\begin{algorithm}[t!]
\caption{Basis Selection Algorithm}
\label{alg:basis}
\begin{algorithmic}[1]
	\State Perform orthogonal-triangular decomposition of $V_{\text{bus}}^K$;
	\State Sort diagonal elements of the upper triangular matrix;
	\State Choose the first $R$ elements that exceed a sufficiently small threshold;
	return the corresponding elements in the permutation matrix
	as indices of the linearly independent rows of $V_{\text{bus}}^K$; 
\end{algorithmic}
\end{algorithm}
\section{Timely Detection and Localization of Events}\label{sec:eventdetect}

Several types of power system events, 
such as switching actions, tap operations, arc and ground faults
can change the effective admittance between the nodes,
thereby resulting in a different admittance matrix.
In this section, 
we propose an online algorithm for tracking changes 
in the admittance matrix of a distribution system
and identifying the events that induced these changes.
This algorithm requires a small amount of data and has a low false alarm rate,
enabling operators to take remedial actions in quasi real-time.

Consider an affine parameterization of the admittance matrix, 
denoted by $Y_{\text{bus}}^{\delta(k)}$, 
where $$\delta(k)=\left\{  
\begin{tabular}{c c}
$0$, & $k  < t$ \\
$1$, & $k \geq t$
\end{tabular}\right.$$ is the discrete mode and $t$ 
is the time that the event has occurred. 
Our goal is to determine $t$ and find out how the admittance matrix has changed 
by estimating $Y_{\text{bus}}^1-Y_{\text{bus}}^0$
using as few successive voltage and current phasor measurements as possible.
The updated entries of the admittance matrix 
indicate the type and approximate location of the event.
For instance, if two elements of the admittance matrix change in a certain way during an event,
it can be attributed to a switch that was opened while another one was closed.

\subsection{Event Detection}

To detect a change in the admittance matrix, 
we estimate the injected current vector at time $k$ from Ohm's law
using the known admittance matrix, $Y_{\text{bus}}^0$, 
and the measured voltage vector at time $k$.
We then compare the estimated injected current vector $\hat{I}_{\text{bus}}$
with the measured current vector $I_{\text{bus}}$ at time $k$ 
to calculate the prediction error:
\begin{equation}
e(k) = I_{\text{bus}}(k) - \hat{I}_{\text{bus}}(k) = I_{\text{bus}}(k) - Y_{\text{bus}}^0 V_{\text{bus}}(k).  
\end{equation}
The series $e(\cdot)$ is white noise if the admittance matrix does not change; 
this can be verified by the turning point test\footnote{
For a detailed discussion on how network topology errors 
can be detected, the readers can refer to~\cite{Clements88}.}.
When the prediction error $\|e(k)\|$ exceeds a predefined threshold, 
we assert that the admittance matrix has changed at time $k$.

\subsection{Event Localization}

The simplest approach to event localization is to run 
the identification algorithm presented in the previous section
upon detection of an event to update the admittance matrix.
In a large distribution system, 
this requires collecting and processing 
a considerable number of $\mu$PMU samples following the detection, 
implying that the identification task may not be accomplished in a timely manner.
To address this shortcoming, 
we propose an identification algorithm that scales with the size of the network
by taking advantage of the knowledge of the admittance matrix before an event
and the fact that only a small number of its elements will change during the event.

Given that the effective admittance 
between just a small number of nodes is expected to change in an event,
the difference between the two admittance matrices corresponding 
to the systems before and after the event must be sparse.
We leverage this sparsity to recover the new admittance matrix:
\begin{align}\label{eq:ydiff}
\min_{Y_{\text{bus}}^1} \| \text{vec}(Y_{\text{bus}}^1 - Y_{\text{bus}}^0)\|_0 &\\
\nonumber \text{subject to} \quad\qquad &I_{\text{bus}}^{t\rightarrow t+K} = Y_{\text{bus}}^1 V_{\text{bus}}^{t\rightarrow t+K}\\ 
\nonumber &Y_{\text{bus}}^1 \in \mathbb{S}^{\sum_{n \in \mathcal N} | \mathcal{P}_n |\times \sum_{n \in \mathcal N} | \mathcal{P}_n |},
\end{align}
in which $Y_{\text{bus}}^0$ is known, $t$ is the time slot when the event is detected, and
\begin{align*}
I_{\text{bus}}^{t\rightarrow t+K}&=\begin{bmatrix}
I_1(t) &\dots &I_1(t+K)\\
I_2(t) &\dots &I_2(t+K)\\
\vdots &\ddots &\vdots \\
I_N(t) &\dots &I_N(t+K)
\end{bmatrix},
\end{align*}
\begin{align*}
V_{\text{bus}}^{t\rightarrow t+K}&=\begin{bmatrix}
V_1(t) &\dots &V_1(t+K)\\
V_2(t) &\dots &V_2(t+K)\\
\vdots &\ddots &\vdots \\
V_N(t) &\dots &V_N(t+K)
\end{bmatrix}.
\end{align*}
It can be readily seen that $\Delta Y\triangleq Y_{\text{bus}}^1 - Y_{\text{bus}}^0$ 
is a symmetric complex matrix as it is the difference of two symmetric complex matrices.
Hence, we have:
\begin{align}\label{eq:Ziso}
&\min_{\Delta Y\in \mathbb{S}^{\sum_{n \in \mathcal N} | \mathcal{P}_n |\times \sum_{n \in \mathcal N} | \mathcal{P}_n |}}\|\text{vec}(\Delta Y)\|_0 \\
\nonumber &\text{subject to} \qquad I_{\text{bus}}^{t\rightarrow t+K} - Y_{\text{bus}}^0 V_{\text{bus}}^{t\rightarrow t+K}
= \Delta Y V_{\text{bus}}^{t\rightarrow t+K},
\end{align}
which can be relaxed and converted 
to the following weighted regularized $\ell_1$-norm optimization:
\begin{equation}\label{eq:Ziso-relaxed}
\text{vec}(\Delta \widehat{Y}) = 
Q_{\Delta Y} \times \arg\min\left\|A x - b\right\|_2^2 + \lambda \sum_i w_i|x_i|,
\end{equation}
where $A=\left({V_{\text{bus}}^{t\rightarrow t+K}}^\top \otimes \mathbb{1}^N\right)Q_{\Delta Y}$, $b=\text{vec}(I_{\text{bus}}^{t\rightarrow t+K} - Y_{\text{bus}}^0 V_{\text{bus}}^{t\rightarrow t+K})$,
and $w_i$ is the weight of the $\ell_1$ penalty which is defined earlier.
This problem is convex and can be solved efficiently with 
only a small number of $\mu$PMU samples compared to the original identification algorithm.
The following proposition gives the necessary and sufficient condition 
for the solution of the $\ell_0$ minimization 
to converge to the true sparse difference matrix
by establishing a minimum bound on the number of $\mu$PMU samples that are required.
\begin{proposition}[From~\cite{Donoho03}]
For any vector $z$, there exists a unique signal $w$ 
such that $z = \Phi w$ with $\| w\|_0 = S$
if and only if $\textbf{Spark}\!\left(\Phi\right) > 2S$.
\end{proposition}
Following this, the proposed event localization algorithm needs
as many $\mu$PMU samples as required for $\textbf{Spark}\!\left(A\right)$ 
to exceed twice the number of elements of the admittance matrix 
that will change during an event.
\section{Performance Evaluation}\label{sec:results}

\begin{figure}[t!]
  \centering
  \includegraphics[width=.45\textwidth]{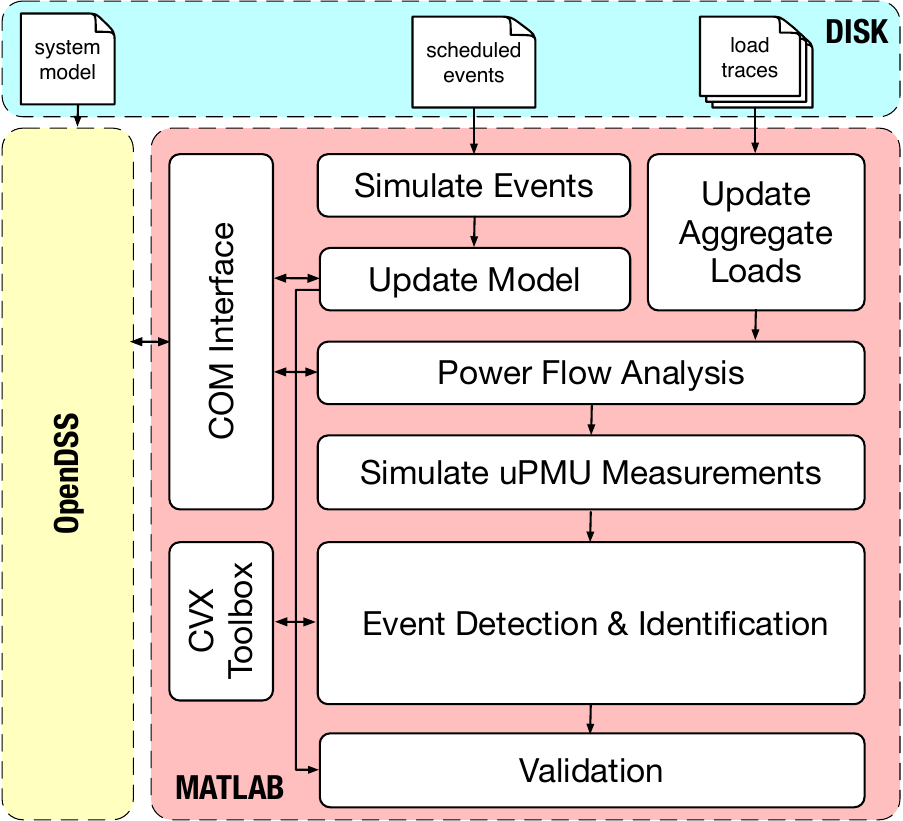}
  \caption{A block diagram of the integrated simulation framework.}
  \label{fig:steps}
\end{figure}

We evaluate the efficacy of the proposed algorithms 
in estimating the model parameters and tracking topology changes 
through power flow studies on test distribution systems under various loading conditions.
To carry out this evaluation, we develop a simulation framework in \textsc{Matlab} 
which integrates built-in graphics and advanced analysis capabilities
with the CVX toolbox for convex optimization~\cite{cvx},
and the Open Distribution System Simulator (\textsc{OpenDSS})~\cite{OpenDSS} for power flow analysis.
The \textsc{OpenDSS} can be controlled from \textsc{Matlab} through a COM interface,
allowing us to load a distribution system model, change its parameters,
perform power flow calculations, and retrieve the results\footnote{The control mode 
is disabled in \textsc{OpenDSS} to ensure that 
the transformer taps are not automatically adjusted during a simulation.
This guarantees that the admittance matrix does not change unless we trigger an event.}.
Figure~\ref{fig:steps} depicts the principal components of this framework.

The simulator takes as input a distribution system model, 
the demand profile of a certain number of homes (for a fixed number of time slots), 
the point of connection of each home (i.e., the node that it is connected to),
and a set of events that should be simulated at specified times.
These events change the admittance matrix in a deterministic way and we record all these changes.
Given this input data, the aggregate demand is calculated at each node
and a power flow study is performed in a loop for each time slot
to compute the voltage magnitude and phase angle at each node,
which are treated as $\mu$PMU measurements.
The proposed convex programs are then solved to 
recover the original admittance matrix from the available $\mu$PMU data
and update it after detecting each event.
The sequence of recovered admittance matrices are eventually compared against 
the sequence of true admittance matrices to quantify the estimation error.
We describe our test cases next.

\subsection{Distribution Feeders}\label{sec:result-feeders}

We evaluate our algorithms on four \textsc{IEEE} test feeders, 
namely 13-bus, 34-bus, 37-bus, and 123-bus systems~\cite{Kersting01};
these unbalanced radial systems operate at a nominal voltage 
of 4.16~kV, 24.9~kV, 4.8~kV, and 4.16~kV, respectively, 
and differ in size and sparsity as shown in Table~\ref{tab:properties}.
The columns of this table respectively represent the test feeder,
the number of nodes in its \textsc{OpenDSS} model,
the rank of $V_{\text{bus}}^K$ (when $K$ is much larger than the number of nodes),
the percentage of $Y_\text{bus}$ elements that are zero (i.e., the sparsity level),
and the absolute value of the largest element of the admittance matrix.
The following observations can be made for each test feeder: 
a) $V_{\text{bus}}^K$ is rank deficient 
which implies that it is impossible to recover 
the full admittance matrix from the $\mu$PMU data,
b) the admittance matrix is extremely sparse,
and c) the admittance matrix has at least one element\footnote{Such elements 
typically correspond to the admittance of switches and voltage regulators.}
that is several orders of magnitude larger than other nonzero elements,
hinting at the possibility that the lasso produces biased estimates for these large elements.

\begin{table}[t!]
    \centering
    \caption{Properties of the radial test feeders.\vspace{-0.2cm}}
    \begin{tabular}{ |l|c|c|c|c| }
     \hline 
     feeder & no. nodes & $\text{rank}(V_{\text{bus}}^K)$ & sparsity level & $|\text{max}(Y_\text{bus})|$\\  [0.75ex]
     \hline
     13-bus  & 35  & 27  & 81.63\% & $10^7$   \\ \hline  
     34-bus  & 95  & 84  & 91.72\% & 402.1    \\ \hline
     37-bus  & 117 & 109 & 92.47\% & 1012.5   \\ \hline
     123-bus & 275 & 254 & 97.39\% & $10^6$   \\
     \hline
    \end{tabular}
    \label{tab:properties}
\end{table}

\begin{figure}[t!]
  \centering
  \includegraphics[width=.40\textwidth]{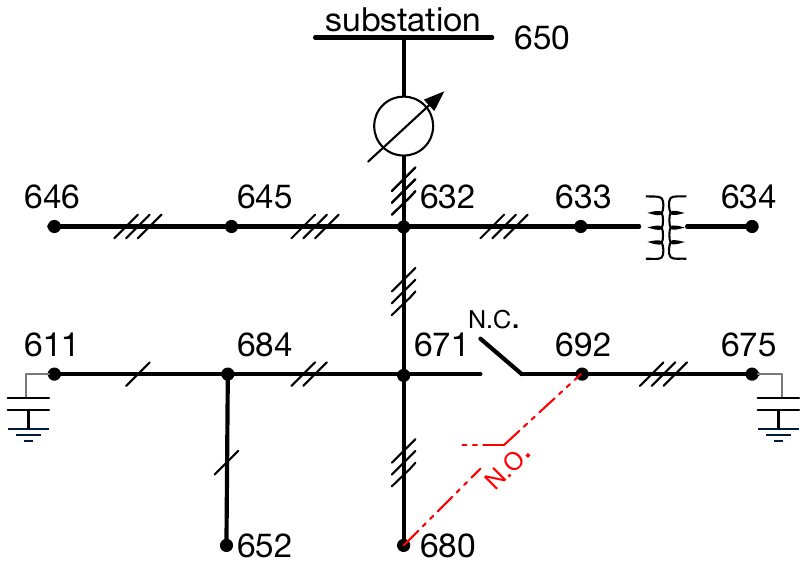}
  \caption{A one-line diagram of the modified IEEE 13-bus test feeder. 
  The dashed line represents the switch added to the original feeder.
  The number of phases connecting two nodes is shown by slashes on the lines.}
  \label{fig:13nodes}
\end{figure}

Moreover, two of these test feeders contain switches
which can be operated to induce a change in the admittance matrix.
The 123-bus test system contains 12 switches that can be operated 
in a certain way to change the topology 
while maintaining its radial structure. 
Hence, it provides an ideal setting for validating 
the event detection and localization algorithm.
Similarly, the 13-bus feeder has a normally closed switch.
We modify this system by adding a normally open switch 
with the exact same configuration 
between Bus~680 and Bus~692 as shown in Figure~\ref{fig:13nodes}.
This creates two feasible radial structures that span all nodes.
We use this feeder to validate both event detection and system identification algorithms. 
The 34-bus and 37-bus feeders do not have a switch 
and are merely used for the purpose of validating the identification algorithm.

\subsection{Residential Loads \& Phasor Measurements}

A node in the distribution system model represents an aggregation point 
where usually a pole-top transformer supplies a small number of residential customers.
Since distribution circuits are not modelled beyond these transformers,
we aggregate demands of downstream customers at the corresponding nodes.
We assume that each node is monitored by a $\mu$PMU 
which measures the magnitude and phase angle of the node voltage and 
the current drawn by the downstream customers once every time slot.

We use real data from the ADRES dataset~\cite{ADRES} 
to model the customers connected to the 13-bus, 34-bus, and 37-bus test feeders.
This dataset consists of high-resolution (1 second) measurements 
of real power, reactive power, and per-phase voltage values 
of 30 Austrian households over 14 days.
To obtain a sufficient number of customers for our simulation, 
we treat a 3-phase load as three separate single-phase loads,
and split 14 days of available data for each household 
into 14 individual loads each representing the demand of a customer over a particular day.
We connect a random number of customers between 5 and 15 to each node,
except the nodes that are terminals of voltage regulators and switches;
no load is connected to these nodes.

In the case of the IEEE 123-bus feeder, 
we need high-resolution measurements of many households, which we lack.
To address this problem, we synthesize residential loads (real power)
using the continuous-time Markov models derived from fine-grained measurements of 
the power consumption of 20 households in Ontario~\cite{Ardakanian11}.
We connect a random number of customers between 5 and 10 to each node, 
except the nodes that represent the terminals of voltage regulators and switches.
Hence, the load distribution is nonuniform across different phases of a node.
We consider a constant power factor of 95\% at each node, 
which is typical for residential loads, 
and set the reactive power accordingly in each time slot.

\subsection{Results}

To deal with the low rank structure of the test feeders, 
we utilize the algorithm proposed in Section~\ref{sec:lowrank} 
to identify the largest part of the admittance matrix 
that could be possibly recovered from the available data.
We also utilize the event detection and localization algorithm
proposed in Section~\ref{sec:eventdetect} 
to track how the admittance matrix changes in an event.
To validate these algorithms,
we perform several simulation runs for each test feeder
where each simulation spans one day divided into 1-second time slots.
In each case, we estimate the admittance matrix 
using both the lasso and the adaptive lasso penalties 
and compare their performance.
We use two metrics to quantify the error incurred in estimating $\mathbb{Y}_{2,2}$:
\begin{enumerate}
    \item[] Metric 1 (M1): $\quad\|\text{vec}(\mathbb{\widehat{Y}}_{2,2}-\mathbb{Y}_{2,2})\|_1$
    \item[] Metric 2 (M2): $\quad\|\mathbb{\widehat{Y}}_{2,2}-\mathbb{Y}_{2,2}\|_F$
\end{enumerate}

To tune the parameters of the convex programs, we search through a reasonable set of values.
Specifically, $\gamma$ and $\lambda$ of the adaptive lasso are chosen from $\{0.5, 1, 2\}$ and 
the set of logarithmically spaced points between $10^{5}$ and $10^{-5}$, respectively.
The same set of values are considered to determine $\lambda$ of the lasso.
Moreover, the OLS estimator is adopted as the weight of the $\ell_1$ penalty in the adaptive lasso.

\begin{table}[t!]
    \centering
    \caption{Estimation error of $\mathbb{Y}_{2,2}$ by the lasso and adaptive lasso 
    in different test feeders.\vspace{-0.2cm}}
    \begin{tabular}{ |c|P{1cm}|P{1cm}|P{1cm}|P{1cm}| }
     \hline 
     \multirow{2}*{feeder} & \multicolumn{2}{c|}{\textbf{lasso}} & \multicolumn{2}{c|}{\textbf{adaptive lasso}} \\ \cline{2-5}
                & M1 & M2 & M1 & M2 \\ \hline
     13-bus     & 4.70   & 4.71   & 1.69 & 1.77   \\ \hline
     34-bus     & 240.51 & 602.89 & 3.08 & 0.37  \\ \hline
     37-bus     & 28.70  & 65.48  & 2.71 & 2.67  \\ \hline
    \end{tabular}
    \label{tab:results}
\end{table}


\subsubsection{Identification with Precise Measurements}
We first explore the scenario in which $\mu$PMU measurements are not affected by noise.
Table~\ref{tab:results} shows the two error metrics when 
the lasso or the adaptive lasso penalty 
is used to identify the admittance matrix of the three test feeders.
In the case of the 13-bus feeder, 
both methods can estimate all elements of $\mathbb{\widehat{Y}}_{2,2}$ except one,
with less than 1\% relative error.
The one element that is not accurately identified corresponds to the substation.
Nevertheless, the adaptive lasso estimates that element 
with a lower error compared to the lasso as suggest by the two error metrics 
representing the overall accuracy of these methods.
Should we leverage the knowledge of the substation type and configuration,
both methods can accurately estimate all elements of $\mathbb{\widehat{Y}}_{2,2}$ ($\text{M1}, \text{M2}<0.01$).

In the case of the 34-bus and 37-bus feeders, 
the lasso fails to accurately identify more 
than one element of the admittance matrix ($\text{M2}\gg\text{M1}$),
resulting in estimation errors which are remarkably higher than the previous case.
We verify that those elements have large admittance values.
However, the adaptive lasso successfully estimates those elements in both feeders,
resulting in relatively smaller estimation errors.

\subsubsection{Identification with Noisy Measurements}
\begin{figure}[t!]
  \centering
  \includegraphics[width=.48\textwidth]{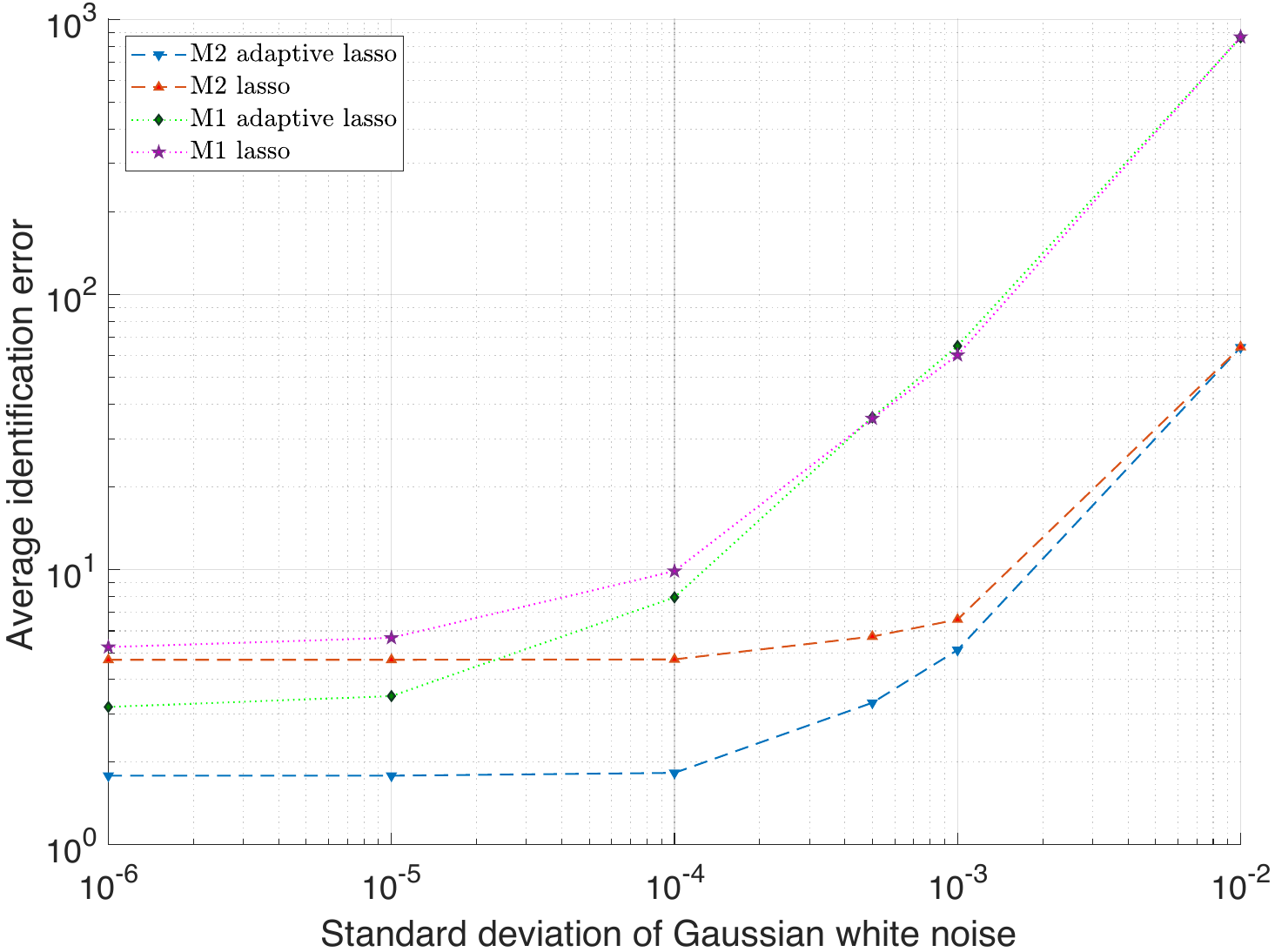}
  \caption{Identification error of the lasso and the adaptive lasso for different noise levels in the IEEE~13-bus test feeder.}
  \label{fig:noisyIDerror}
\end{figure}

We now explore the scenario in which $\mu$PMU measurements are noisy.
We only show the results for the 13-bus feeder due to space limitations.
To simulate the measurement error, 
we add a white Gaussian noise with $\sigma^2$ variance to node voltages
and treat them as $\mu$PMU measurements\footnote{We have observed that perturbing
node voltages by a small Gaussian noise does not completely eliminate the low rank problem.}.
We try different values of $\sigma$ and perform ten simulation runs for each value.
In particular, we increase the standard deviation from $10^{-6}$ to $10^{-2}$
and report the mean value of the two error metrics over these runs.
Figure~\ref{fig:noisyIDerror} shows the mean estimation error of the lasso and the adaptive lasso.
It can be seen that both methods are quite sensitive to noise. 
When $\sigma=10^{-6}$, the estimation error of both methods 
is quite similar to the scenario with precise measurements (see Table~\ref{tab:results}).
The adaptive lasso can accurately identify all elements of $\mathbb{\widehat{Y}}_{2,2}$, 
except for one element (as discussed earlier) as long as $\sigma\leq 10^{-3}$. 
Should $\sigma$ exceeds this level, 
the adaptive lasso fails to identify several elements and both error metrics increase significantly.
Unlike the adaptive lasso, the lasso is less robust to noise and only
yields sufficiently accurate estimates when $\sigma\leq 10^{-4}$.


\subsubsection{Event Detection and Localization}
We finally validate the proposed event detection and localization algorithm
by simulating a line tripping event and a switching operation in the 13-bus test feeder,
and a switching operation in the 123-bus test feeder.
These events are triggered after the admittance matrix has been identified 
for the initial configuration of the distribution system.

\begin{figure}[t!]
  \centering
  \includegraphics[width=.48\textwidth]{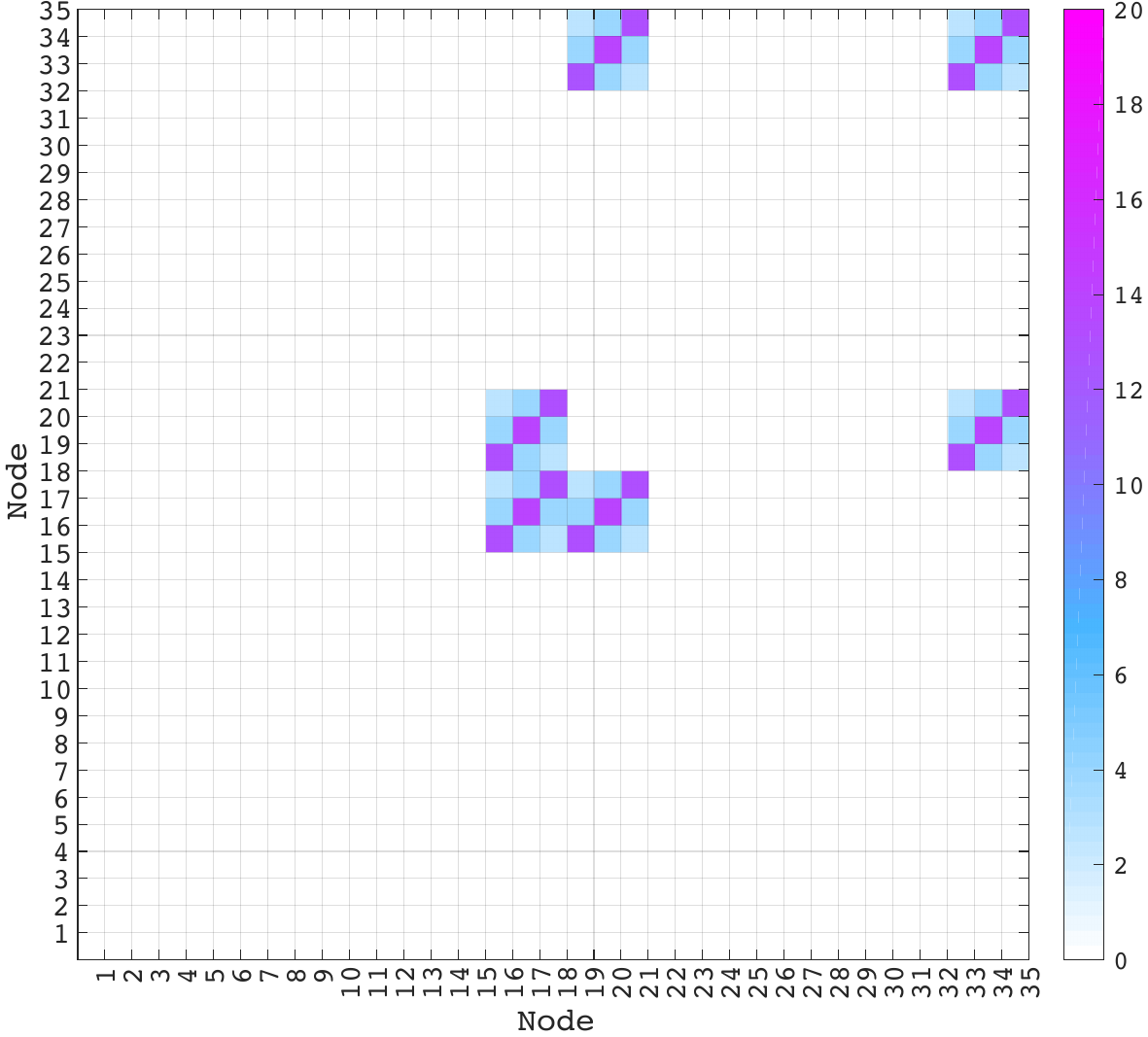}
  \caption{A colored representation of $\Delta\widehat{Y}$
  when 23 $\mu$PMU samples are used to recover the admittance matrix following the detection of the switching event.}
  \label{fig:heatmap1}
\end{figure}

We first focus on the 13-bus test feeder.
We introduce a line tripping event by disconnecting 
the single-phase line between Bus 611 and Bus 684,
and a switching event by closing the switch between Bus 692 and Bus 680
while opening the switch between Bus 671 and Bus 692.
Both events will change the admittance matrix 
and therefore must be identifiable from the available data.
We observe that the proposed algorithm detects the event in both cases 
in the same time slot that it occurs,
i.e., immediately after processing the $\mu$PMU data for that time slot.
For the switching event, $\Delta\widehat{Y}$ can be estimated 
with relatively high accuracy ($\text{M1}=0.17,\text{M2}=0.10$)
using 23 $\mu$PMU samples following the detection of this event,
as shown in Figure~\ref{fig:heatmap1}.
In this figure, the color of a cell located at row $i$ and column $j$
represents the value of $|\widehat{Y}^1(i,j)-\widehat{Y}^0(i,j)|$.
It can be readily seen that all elements of $\Delta\widehat{Y}$ are zero 
except for 6 three by three submatrices which have changed due to this event,
enabling us to locate the event within a small geographical area.
We verified that these submatrices correspond to 
the admittance of the two switches that are operated.
Note that these switches are located on three-phase lines.
The estimation error will increase significantly 
if we use fewer samples to identify $\Delta\widehat{Y}$.

Turning our attention to the 123-bus test feeder, 
we simulate a line switching event by closing 
the switch between Bus 13 and Bus 152 
while opening the switch between Bus 151 and Bus 300 
(refer to~\cite{testfeeders} for the topology of this feeder).
The simultaneous switching operation
maintains the radial structure of this distribution system.
The proposed algorithm detects the event in the same time slot that it occurs
and accurately identifies $\Delta\widehat{Y}$ ($\text{M1}, \text{M2}<0.1$)
using a small number of $\mu$PMU samples following the detection.

\section{Conclusions}

Widespread adoption of distributed energy resources in power distribution grids
calls for an advanced operation paradigm centered around monitoring, diagnostics, and control.
While the knowledge of the distribution system model is crucial 
for most diagnostics and control applications, 
this model is often unavailable or outdated in practice.
This paper studies how the admittance matrix of a power distribution system 
can be identified from limited voltage and current phasor measurements of inexpensive distribution PMUs.
It proposes tractable convex programs 
to recover the admittance matrix in various settings 
and to track changes in the network topology,
and investigates the fundamental limitations of these techniques.
There are several avenues for future work.
The $\mu$PMU installation is currently limited in low-voltage distribution networks.
To address this problem, we intend to develop techniques
that leverage both smart meter and $\mu$PMU data 
to improve identifiability of low-voltage distribution grids.
We also plan to develop an identification algorithm 
that can deal with hidden states in the network.
Furthermore, the proposed technique can be computationally expensive
when it is applied to a distribution network with thousands of nodes.
We intend to develop a distributed identification algorithm 
for large distribution networks.

\bibliographystyle{IEEEtran}
\bibliography{mybib}
\end{document}